\documentclass[11pt]{article}
\usepackage{eamt24}
\usepackage{times}
\usepackage{url}
\usepackage{latexsym}
\usepackage[small,bf]{caption} 
\usepackage{amsmath,amssymb}

\newcommand{\confname}{EAMT 2024}

\newcommand{\conffilename}{eamt24}
\newcommand{\downloadsite}{\url{http://eamt2024.sheffield.ac.uk}}

\newcommand{\researchtechlength}{$10$ (ten)}
\newcommand{\researchtransuserslength}{$10$ (ten)}
\newcommand{\implementationslength}{$6$ (six)}
\newcommand{\projectlength}{$2$ (two)}

\title{Using Machine Translation to Augment Multilingual Classification}

\author{Adam King\\
  GumGum\\
  {\tt aking@gumgum.com}
  }

\date{May 6, 2024}

\begin{document}
\maketitle
\begin{abstract}
An all-too-present bottleneck for text classification model development is the need to annotate training data and this need is multiplied for multilingual classifiers. Fortunately, contemporary machine translation models are both easily accessible and have dependable translation quality, making it possible to translate labeled training data from one language into another. Here, we explore the effects of using machine translation to fine-tune a multilingual model for a classification task across multiple languages. We also investigate the benefits of using a novel technique, originally proposed in the field of image captioning, to account for potential negative effects of tuning models on translated data. We show that translated data are of sufficient quality to tune multilingual classifiers and that this novel loss technique is able to offer some improvement over models tuned without it.

\end{abstract}

\section{Introduction}
One of the most common uses of machine learning for natural language processing (NLP) is the classification of text into one of multiple mutually-inclusive or mutually-exclusive labels. Recently, generative LLMs, such as PaLM \cite{flan} and ChatGPT \cite{chatgpt} have shown exciting and impressive capabilities to do zero- or few-shot prompting, classify text given only a few examples for the task across a variety of languages. Nevertheless, it is still the case that the highest performing and most efficient means to classify text is the use of a bespoke classifier trained with hundreds or thousands labeled examples \cite{pires2019}, particularly when the task requires a level of human-like subjectivity or general reasoning ability \cite[see discussion]{kocon2023}. To this end, finding or creating a corpus of labeled examples is a necessary step in the creation of any classifier. 

For high-resource languages like English, which have many existing labeled corpora available and large populations of annotators on crowd-sourced workers such as Amazon Mechanical Turk, the challenge of creating or finding training and evaluation data can be costly, but not prohibitively so. Yet, for lower-resourced languages which lack existing annotated corpora and have smaller or even non-existent populations on these large annotation platforms, acquiring the required training data can prove to be much more difficult. Moreover, if the model is intended to be able to perform the same classification across multiple languages, the time and effort required to annotate training data becomes multiplicative. Fortunately, classification is not alone in the applications of machine learning in NLP. Machine translation (MT) has seen major improvements in recent years \cite{stahlberg2020}, accelerated by the adoption of the transformer architecture \cite{vaswani2017}. 

To date, several options for high quality machine translation currently exist, between API services and open-source models. MT API services, such as Google translate, have become nearly ubiquitous, provide high quality translations, while still being relatively inexpensive. In fact, in one experiment, translating data using Google translate into English and using existing English-trained classifier models outperformed certain models trained on the original language directly \cite{araujo2016}. In addition to MT API services, several open-source translation models are easily available, such as the multilingual M2M100 model \cite{m2m100}, NLLB200 model \cite{nllb200} or the over 1400 models trained by the University of Helsinki \cite{tiedemann2020}, with many of these models have performance that approaches or exceeds that of MT APIs \cite{stahlberg2020}. 


With this in mind, it may be the case that translating an existing, labeled dataset with one of the aforementioned MT options is a feasible alternative to creating a novel dataset directly in that language. This has several benefits. Firstly, it avoids the problem of existing corpora or annotation options not existing for the language in question. Secondly, it minimizes  the data needed for multilingual models and allows annotations for one language to serve another. Here, we ask if it is possible to use MT to train a multilingual model, given only original, annotated data for a single language.

Of course, the potential benefits of using MT to train a multilingual model are still affected by the old machine learning adage: garbage in, garbage out. Even the best translations, either human or machine, will lose some of the information of the original language, which will inevitably lead to dropped performance for a model trained on the translated examples. Fortunately, the problem of training models using semantically similar but imperfect pairs of data is not unique to the task at hand and there is a growing body of research which may provide some benefit. In particular, image captioning is a task to generate the ideal natural language text caption for an image  and these captioning models must learn to represent semantically related data from very different modalities similarly, i.e., text and images \cite{li2021}. In this way, image captioning is somewhat analogous to the task of training on translated data, where we want to have semantically identical text from different languages predicted to have the same labels. As a result, we ask in addition whether some of the model training techniques used in image captioning models can lead to improved performance for multilingual models trained using MT data.

\section{Related Work}


This work is by no means the first to suggest the usage of machine translation to create or augment datasets for lower resourced languages.  Wei and Pal \shortcite{wei2010} and Pan et al. \shortcite{pan2011} augmented Chinese language corpora with annotated data translated from English to improve the performance of a Chinese-language sentiment analysis model. On the other hand, Barriere and Balahur \shortcite{barriere2020} and Ghafoor et al. \shortcite{ghafoor2021} used existing API translation services to translate annotated data from English into lower-resourced languages and trained classifiers solely on these translated data, finding that classifiers trained on translated data were fairly accurate but did see drops in performance, likely due to the effects of imperfect translations of the training data. 

It should be noted that training a model from scratch is not the only means to create an accurate classifier, particularly for lower-resourced languages. Large multilingual transformer models such as {\sc m-BERT} \cite{bert}, {\sc xlm-RoBERTa} \cite{xlm-roberta} or {\sc gpt-3} \cite{gpt3} have been shown to have the ability to generalize from one language to the other, i.e., train in one language and improve test performance in another language, \cite{pires2019}, but benefits of this vary on the languages in question, with languages that share closer genealogical origin or structural similarities benefiting more from inter-language transfer. Regardless, training a model with examples of a particular language dependably yields the best classifier for new data in that language. 

Nevertheless, to date there has been no investigation of how fine-tuning large multilingual transformer models on translated data affects final performance compared to simple interlanguage transfer. Moreover, previous work to train models using translated data employed a naive approach, treating translated data as if it were no different than original, untranslated data which annotated itself. In this work, we investigate both how multilingual transformer models trained on translated data perform compared to interlanguage transfer and explore a means to mitigate imperfect translation quality when creating these training datasets.

\section{Image captioning and Image-Text Contrastive Loss}
Image-text Contrastive (ITC) loss is a technique used when training multimodal models to caption images with natural language descriptions \cite{li2021}. For example, {\sc BLIP} \cite{blip} is a image-captioning model that was trained with a mix of human- and artificially-annotated images where ITC loss was integral to the models ability to learn from noisy, artificially-annotated data. ITC loss, then, has been shown to mitigate negative effects of both noise and different modalities for multimodal models.

At an intuitional level, these captioning models decompose text and images into a shared embedding space and ITC loss seeks to penalize cases where related image-text pairs are dissimilar in this shared embedding space. In other words, ITC looks seeks to bring semantically related items from disparate modalities closer in a shared embedded space and has empirically improved image-captioning models, with little impact on training time or resources. 

Training multilingual classification models with translated data bears a similarity to captioning, though rather than have semantically related examples from different modalities, there are semantically parallel data in different languages. That being the case, we will be a slightly modified form of ITC loss, namely original-translated contrastive (OTC) loss, to enforce similarity within a batch between data from the original language and its translated counterpart. Like ITC loss, OTC loss penalizes a transformer model for dissimilar embedding representations for translated pairs. One way to think of it is that this loss encourages the model to embed sentences with the same meaning identically, regardless of language.

In detail, we implement OTC loss as follows. We begin by deriving a probability of each original/translated pairing in a training minibatch, $p^{o2t}$ and $p^{t2o}$, that is, which original examples pairs with which translated example and vice versa.

\begin{equation}
    p^{o2t}_m = \frac{exp(s(O,T_{m})/\tau)}{\Sigma_{m}^{M} exp(s(O,T_{m})/\tau)}
\end{equation}
\begin{equation}
    p^{t2o}_m = \frac{exp(s(T,O_{m})/\tau)}{\Sigma_{m}^{M} exp(s(T,O_{m})/\tau)}
\end{equation}

Here, $s(T,O)$ is a similarity function between the original, untranslated data and the translated examples in a minibatch. We compute $s(T,O)$ by first extracting and normalizing the embedding for the initial {\sc [cls]} token after the final attention head of the encoder stack in {\sc m-BERT}, computing a pairwise dot product for all possible pairs of original and translated data and dividing by $\tau$, which is a learnable parameter. We then apply the softmax function as a way to represent the likelihood of each original/translated match. Ideally, each correct original/translated pair will have the most similar embeddings, resulting in a value close to 1 after softmax. As a final step, we compute the cross-entropy between the result of the previous step and a target vector which encodes the correct original/translated pairs, weighting this by a hyperparameter, $\alpha_{otc}$. Following {\sc BLIP} \cite{blip}, we set $\alpha_{otc}=.4$ for all runs. 

\begin{multline}
    \ell_{otc} = \alpha_{otc} * \frac{1}{2} \mathbb{E}_{(O,T)}[H(\textbf{y}^{o2t}(O), \textbf{p}^{o2t}(O) + \\
    H(\textbf{y}^{t2o}(T), \textbf{p}^{t2o}(T)]
\end{multline}

\section{Experiments}
\subsection{Data}
For these experiments, we use a multilingual dataset of Amazon product reviews across 6 languages: English, Spanish, French, German, Chinese and Japanese \cite{amazon_multi_reviews}. This dataset is comprised of over 1 million total examples, split into a train and test partition. The reviews are equally distributed across the six languages, as well as the total stars given to the reviewed product (1-5) for both the train and test partition, i.e., each number of stars comprises 20\% of the examples for that language. This dataset is particularly useful due to its size, number of available languages and presence of an established training and test data split.

We began by translating each review from the training partition of the original dataset into each of the other respective languages and assigned the same star value to the review (see example \ref{table:translated-data}), i.e., if a review was originally in English and had star star, when translating it into French it would also be labeled with one star. We did this translation once before carrying out the rest of the experiment to ensure each classifier would be trained on the same set of translations. To translate, we used a single multilingual translation model, M2M100 \cite{m2m100}. We chose to use a single multilingual translation model in order to mitigate any potential differences from translation quality coming from different machine translation architectures.

\begin{figure*}
\begin{center}
\begin{tabular}{c|c|c|l|c}

id & translated & language & text & stars \\
\hline \hline
1 & 0 & en & My daughter really likes the backpack and ... & 5 \\
1 & 1 & es & Mi hija realmente le gusta el bolsillo y ...	 & 5 \\
... & ... & ... & ... & ... \\
2 & 0 & en & This product is BS, I washed my face with hot water ... & 1 \\
2 & 1 & fr & Ce produit est BS, je me suis lavé le visage à l'eau chaude ... & 1 \\
... & ... & ... & ... & ... \\
\end{tabular}
\end{center}

\caption{Example original and translated data. Each unique review (id) in the original dataset was translated to the other languages and assigned the same star value. Texts truncated here for formatting.} \label{table:translated-data}
\end{figure*}

\subsection{Experiment design}
To investigate any potential improvement in classifier accuracy with the use OTC loss, we fine-tuned pretrained transformer models on datasets that included original, untranslated data for a single language\footnote{We restricted the experimental conditions to only including a single language's original data, rather than use the full set of $6!=720$ possible permutations of language combinations for the sake of efficiency and resources.} and only translated data for all others in the six language set. As an example, in one training run, the model would be tuned on the original English training data and only translated data for all other languages, which were translated from the set of the original English data. We did this for all six languages in the original set to ensure any results were not restricted to one language in the dataset. Though the exact training examples varied for each model, we tested each on the original testing split of the dataset, which was solely comprised of original data, i.e., non-translated, for the six languages. 

In each case, we tuned a multilingual {\sc DistilBERT} model \cite{distilbert}, a distilled version of the original multilingual {\sc m-BERT} \cite{bert}, to predict the number of stars on a review as a categorical classification problem, using categorical cross-entropy loss and varying between using OTC loss as an additional loss parameter between runs. We chose to use a distilled variant of {\sc BERT} due to the distilled variants increased speed of training, while still maintaining 97\% of overall language understanding of the original. 

Because of the mechanics of OTC loss, each translated datum must have an original match in the minibatch and each original must have at least one translated variant. As such, we constructed minibatches during training such that half the samples were always original, untranslated data and the other half were a randomly selected translated example for each original datum. For each original example, we randomly selected a translated example from the other languages, meaning that the model saw an equal number of original and translated examples during tuning overall, though it saw far fewer individual examples of each translated language, i.e., roughly $\frac{1}{5}$. For simplicity, we restricted our tests to a 1:1 original:translated ratio and we used the same batch sampling method for runs without OTC loss, to make results more easily comparable.

For each tuning run, we used a batch size of 32 (16 original and 16 translated examples per batch)\footnote{For baseline conditions where there was no translated data, mini-batching happened as normal with 32 examples original, untranslated data per batch.} and used the AdamW \cite{adamw} optimizer with a linear warm-up of 500 updates with a learning rate of 2e-5. All training was done on {\sc g5.2xlarge} AWS instances which contain NVIDIA A10G GPUs. We tuned 3 separate tuning runs for each set of hyperparameters and report their mean values in the next section.

\section{Results}
In these experiments, we asked two simple questions: 1) how feasible is it to tune a multilingual transformer model on translated data and 2) does the inclusion of OTC loss improve model performance for languages where only translated training data was used. 

In answer to the first, for each of the six languages in the original dataset, models fine-tuned with translated data showed higher F1-micro scores\footnote{F1-micro is an example-weighted version of the F1-score, which is the harmonic mean or precision and recall. For more details on F1-score, see \cite{jurafsky2008}.} on the held-out test set, compared to models trained with only original data for a single language (see Table \ref{table:translated-results}). As was expected from Pires et al. \shortcite{pires2019}, even if a model was never exposed to data for a language, original or translated, the final model did have F1-micro greater than chance for that language (which would be 20\% for a balanced, 5-label problem), indicating there was interlingual knowledge transfer happening within the model during training. Moreover, it appears that there was more transfer between related, similar languages, compared to more dissimilar languages; models trained with data for a European language showed higher performance on other European languages, compared to Japanese or Chinese. Nevertheless, for all languages, the use of translated data did show a noticeable improvement (.02-.11), though for each language, models trained with only translated data did underperform models trained with the full set of origina, untranslated training examples for that language (.07-.12).

That said, it is clear that the use of translated training data does improve model performance, even if the trained model only sees translated examples for that language. It should also be noted that due to the batching and sampling strategy used here, models trained with translated data saw far fewer examples of each language where they only saw translated data. That is, because each original review was paired with a single translated example out of five possible translated, these models were exposed to roughly one fifth of the data for translated languages and still saw a sizable boost in performance.

\begin{table}
\begin{center}
\begin{tabular}{c|c|c|c}

Language & \multicolumn{3}{c}{F1-micro} \\
\hline
& No data & Translated & Original \\
\hline \hline
{\sc en} & 0.407 & 0.481 & 0.554\\
{\sc fr} & 0.379 & 0.468 & 0.544\\
{\sc de} & 0.359 & 0.465 & 0.581\\
{\sc es} & 0.376 & 0.474 & 0.55\\
{\sc ja} & 	0.307 & 0.396 & 0.543\\
{\sc zh} & 0.352 & 0.372 & 0.458\\
\end{tabular}
\end{center}

\caption{F1-micro for models trained with no samples for the specified language (No data), with only translated samples (Translated) and with the original training data for that language (Original). All languages saw a sizeable boost to performance over their respective baselines when using translated data (.02-.11) but all languages did perform markedly better when given actual data for each language.} \label{table:translated-results}
\end{table}

Moving on to the effect of OTC loss, Table \ref{table:pairwise-results} shows the mean F1-micro per language in the testing set, for models fine-tuned using original data for the specified language and translations for all other languages. For all languages, models trained using OTC loss saw an improvement over models trained without for all languages except Chinese, which showed a mixed set of negligible differences or lowered performance. However, these values include runs where the specific language was included as original, untranslated data. When averaging across all runs where a language in the testing set was only represented by translated data, OTC loss shows an improvement over models trained without it for all languages. Table \ref{table:avg-results} shows the mean F1-micro for all models trained where the specified language was not the original language.

\begin{table}
\begin{center}
\begin{tabular}{c|c|c}

Language & \multicolumn{2}{c}{F1-micro} \\
\hline
& No OTC & OTC \\
\hline \hline
{\sc en} & 0.479 & {\bf 0.483}\\
{\sc fr} & 0.464 & {\bf 0.472}\\
{\sc de} & 0.463 & {\bf 0.467}\\
{\sc es} & 0.472 & {\bf 0.476}\\
{\sc ja} & 	0.393 &	{\bf 0.399}\\
{\sc zh} & 0.368 &	{\bf 0.376}\\
\end{tabular}
\end{center}

\caption{Comparison on final performance per language for models that only included translated examples for the specified language. Though the gain was less than .1, each language consistently performed better when trained with OTC loss.} \label{table:avg-results}
\end{table}

To ensure that the results here were in fact statistically significant, we fit a linear mixed-effect model to predict final model F1-micro for a language, given the hyperparameters of a particular tuning run. Mixed-effect models are able to accurately evaluate the contribution of different fixed-effect independent variables, e.g., whether OTC was used when training a particular model, on dependent variables, e.g., the final accuracy of the trained model, all the while being robust to expected random variance between trials, e.g., because of random initialization and batching, some deep learning models score higher than others with identical hyperparameters  (see Baayen et al. \shortcite{baayen2008}, Jaeger \shortcite{jaeger2008} for more). 

This statistical model was fit to predict per-language test f1-micro, given a random effect of each model run and three fixed effects: i) the tested language, ii) the identity of the single original language and iii) whether OTC loss was added. OTC was found to have a significant, positive effect ({\sc coef}=0.036, {\sc Std.Error}=0.017, for all model details see \ref{figure:lme-results}), indicating that even after taking into consideration differences between languages and random variance for each multilingual model, the inclusion of OTC loss did yield an improved final model F1-micro. 

\begin{table*}
\begin{center}
\begin{tabular}{c|c|c|c|c|c|c|c}

Orig. Training Language & OTC & {\sc en} & {\sc fr} & {\sc de} & {\sc es} & {\sc ja} & {\sc zh}\\ 
\hline \hline
{\sc en} & No OTC & 0.548 & 0.488 & 0.493 & 0.489 & 0.425 & 0.423\\
 & OTC & {\bf 0.553} & {\bf 0.507} & {\bf 0.522} & {\bf 0.512} & {\bf 0.434} & 0.422\\
\hline
{\sc fr} & No OTC & 0.504 & 0.539 & 0.504 & 0.493 & 0.424 & {\bf 0.426}\\
 & OTC & {\bf 0.512} & 0.539	& {\bf 0.517} & {\bf 0.511}	& {\bf 0.428} & 0.412\\
 \hline
{\sc de} & No OTC & 0.514 & 0.495 & 0.577 & 0.495 & 0.436 & 0.427\\
& OTC & {\bf 0.524} & {\bf 0.506} & {\bf 0.581} & {\bf 0.506} & {\bf 0.449} & 0.425\\
\hline
{\sc es} & No OTC & 0.506 & 0.497 & 0.500 & 0.544 & 0.433 & {\bf 0.419} \\
& OTC &	{\bf 0.523} & {\bf 0.510} & {\bf 0.518} & {\bf 0.548} & {\bf 0.441} & 0.413\\
\hline
{\sc ja} & 	No OTC & 0.470 & 0.460 & 0.477 & 0.468 & {\bf 0.526} & {\bf 0.436}\\
 & OTC &	{\bf 0.493} & {\bf 0.474} & {\bf 0.499} & {\bf 0.487}  & 0.522 & 0.424 \\
 \hline
{\sc zh} & No OTC & 0.486 & 0.439 & 0.441 & 0.444 & 0.398 & 0.482\\
 & OTC & {\bf 0.488} & {\bf 0.467} & {\bf 0.473} & {\bf 0.472} & {\bf 0.421} & {\bf 0.503}\\
\end{tabular}
\end{center}

\caption{F1-micro results on untranslated test data. Each row shows the per-language performance for models trained with original data for the specified language and translated data for all other languages, using OTC loss and without. Each cell shows the mean of 3 runs per condition. Bolded values show a difference of .03 or greater.} \label{table:pairwise-results}
\end{table*}

\section{Discussion and future directions}
We investigated the feasibility of using translated text to fine-tune a multilingual transformer model, as well as any potential gains by utilizing a novel application of deep learning technique to improve performance. We found that models trained using only translated data for a language do show a noticeable improvement over baselines, though as expected, there was still a performance drop from using original, untranslated data for that language. We also found that slight further gains can be achieved by the use of OTC loss, suggesting that training the model in such a way where it is sensitive to potential data issues improves its ability to generalize.

Granted, this is a very open problem and results of using translated data to tune a multilingual classifier will vary highly depending on the quality of MT model used, architecture of the classifier being tuned and the type of classification being modeled. Nevertheless, the results here are exciting for multiple reasons. Firstly, as suggested by previous works \cite[as an example]{shalunts2016}, MT is useful tool for language-specific dataset creation when creating a dataset for that language directly may prove difficult. In this case, we showed that {\sc m-BERT} models tuned on translated examples showed large gains over simple multilingual transfer during training. This is particularly interesting given that for each translated language, the model was only given a fraction of samples compared to the original language due to the 1:1 ratio of original and translated data. A future direction for this work may be to adjust this ratio or the number of languages in the dataset to investigate how this affects model training. Secondly, the use of OTC loss was shown to lead to a small, but robust boost to performance. This suggests that methods of mitigating the natural effects of translation have a potential to bridge the gap, so to speak, between models trained on translated data and on datasets in the target language directly. Particularly relevant, Chinese, which is linguistically dissimilar from the majority of languages in the set used here, showed a mixed ability to benefit from training with other languages, but a clearer improvement using OTC loss. This may suggest that OTC loss is able to mitigate structural differences between languages and a future direction for this may be to explore exactly how OTC loss affects individual examples and how other noise-reduction techniques may lead to further gains in model performance. 

Putting this together, this is an indication that MT-augmented datasets stand as a good first step for developing multilingual classification models. Given that MT can quickly and efficiently expand an annotated dataset from one language into another and that translated dataset is of sufficient quality to improve over basic interlingual transfer, this technique has great potential to expanding classification tasks to new languages quickly. In addition, OTC loss may be able to slightly but significantly increase the quality of these models with no additional data. All in all, we are confident that the use of MT augmentation is an exciting and interesting topic for future exploration.

\section{Acknowledgements}
This work was carried out as a part of the R\&D for GumGum's Verity product. Special thanks go to the members of the AI team (names, names, names) for their suggestions and audience during brainstorming, development and analysis of this project.

\newpage
\bibliography{eamt24}
\bibliographystyle{eamt24}

\appendix
\begin{figure*}
\begin{verbatim}
              Mixed Linear Model Regression Results
==========================================================
Model:              MixedLM  Dependent Variable:  test_acc
No. Observations:   108      Method:              REML    
No. Groups:         18       Scale:               0.0038  
Min. group size:    6        Log-Likelihood:      111.7634
Max. group size:    6        Converged:           Yes     
Mean group size:    6.0                                   
----------------------------------------------------------
                       Coef.  Std.Err.   z    P>|z| [0.025 0.975]
----------------------------------------------------------
Intercept              0.465    0.024 19.128 0.000  0.418  0.513
otc[T.True]            0.036    0.017  2.105 0.035  0.002  0.069
original_lang[T.en]   -0.020    0.028 -0.714 0.475 -0.074  0.035
original_lang[T.es]   -0.012    0.028 -0.432 0.666 -0.066  0.042
original_lang[T.fr]   -0.015    0.028 -0.555 0.579 -0.070  0.039
original_lang[T.ja]   -0.030    0.028 -1.093 0.274 -0.085  0.024
original_lang[T.zh]   -0.050    0.028 -1.801 0.072 -0.104  0.004
test_lang[T.en]        0.016    0.020  0.762 0.446 -0.025  0.056
test_lang[T.es]        0.003    0.020  0.130 0.897 -0.037  0.043
test_lang[T.fr]       -0.000    0.020 -0.005 0.996 -0.040  0.040
test_lang[T.ja]       -0.061    0.020 -2.972 0.003 -0.101 -0.021
test_lang[T.zh]       -0.068    0.020 -3.336 0.001 -0.108 -0.028
Group Var              0.001    0.009                           
==========================================================
\end{verbatim}

\caption{Full model details for MLE model trained to predict F1-micro per laguage. {\sc otc} has a positive contribution to an increase F1-micro score, even when controlling for variance between languages and model runs.} \label{figure:lme-results}
\end{figure*}

\end{document}